\documentclass[runningheads]{llncs}
\usepackage[T1]{fontenc}
\usepackage{graphicx}
\usepackage{booktabs}
\usepackage[misc]{ifsym}

\usepackage{wrapfig}
\usepackage[disable]{todonotes}
\usepackage{amsmath}
\usepackage{multirow}
\usepackage{hyperref}
\usepackage{mdframed}
\usepackage{float}
\usepackage{subcaption}
\usepackage{orcidlink}
\usepackage[normalem]{ulem}
\usepackage{booktabs}
\usepackage{array}

\begin{document}

\title{Beyond Noise: Understanding and Overcoming Temperature Effects in Analog DNN Inference}

\titlerunning{Beyond Noise: Understanding and Overcoming Temperature Effects}

\author{Niklas Summ\orcidlink{0009-0002-9919-7095} \and
	Xiao Wang\orcidlink{0009-0001-5064-7940} \and 
	Hendrik Borras\orcidlink{0000-0002-2411-2416} \and
	Bernhard Klein\orcidlink{0000-0003-0497-5748} \and
	Holger Fröning\orcidlink{0000-0001-9562-0680}}

\authorrunning{N. Summ et al.}

\institute{Hardware and Artificial Intelligence Lab, Institute of Computer Engineering, Heidelberg University, Germany\\
\email{niklas.summ@stud.uni-heidelberg.de, \{xiao.wang,hendrik.borras,bernhard.klein,holger.froening\}@ziti.uni-heidelberg.de}}

\maketitle

\begin{abstract}
The energy efficiency of analog computing makes it one of the most promising candidates for deploying resource-intensive machine learning workloads on constrained platforms such as mobile and embedded devices.
However, analog accelerators are inherently susceptible to noise and non-idealities arising from physical component variations, whose behavior is further sensitive to environmental factors.
These effects can significantly degrade inference accuracy.
In this work, we conduct a comprehensive experimental study on a representative example of analog hardware to investigate the impact of temperature.
We first characterize the behavior of stochastic and systematic non-idealities across a range of operating temperatures.
Following this, we compare a set of simulation-based and hardware-based mitigation strategies aimed at improving robustness against temperature-induced performance degradation.
Our results suggest that temperature-induced degradation is driven primarily by systematic non-idealities rather than stochastic noise alone. Noise-aware training improves robustness, while hardware-in-the-loop training and temperature-aware calibration provide the strongest accuracy retention across varying thermal conditions.

\end{abstract}

\section{Introduction}
The computational demands of deep neural networks continue to grow rapidly, creating increasing pressure on conventional digital hardware. 
While modern GPUs and neural processing units provide high throughput, their energy consumption and data movement overhead remain major obstacles.
Analog computing has therefore re-emerged as a promising alternative for neural network inference~\cite{9197673,bss2-figs}.
It exploits physical quantities without digitization to perform highly-parallel energy-efficient multiply-accumulate operations.

However, the benefits of analog computing come at the cost of reduced computational reliability. 
In contrast to digital arithmetic, analog computation is inherently affected by non-idealities such as device-to-device variations, drift, thermal fluctuations, nonlinearities, and noise. 
In the following, we will use the term ``noise'' to refer to stochastic effects, while the term ``systematic non-idealities'' will refer to contributions that offset the signal amplitude in a predictable manner.
While there are recent works which use noise as a computational resource~\cite{fbp2024,fbp2025}, we note that even in such cases, noise can still degrade performance.
Thus, robust deployment of neural networks on analog hardware requires characterization and training methods that explicitly account for imperfect and changing computation conditions.
Of particular interest for edge deployments, such as personal computing, are temperature variations, as these cannot reasonably be kept constant.

A widely used approach to improve robustness against analog computation errors is noisy training~\cite{noisy_machine,bernhard-incremental}, where noise is injected during training to expose the model to perturbations similar to those expected during inference. 
Prior work has shown that noisy training can substantially improve robustness compared to na\"ive training, quantization-aware training, or general robustness methods such as Sharpness-Aware Minimization~\cite{accml2025}.

From a simulation perspective, recent work identifies noisy training as the strongest baseline for robustness under noisy analog computations, but also as fundamentally limited when the noise level during deployment differs from the noise level assumed during training.
Variance-Aware Noisy Training (VANT)~\cite{xiaowang2025} addresses this limitation by exposing the model to a range of possible inference-time noise conditions rather than a single fixed noise configuration.

However, these studies generally rely on abstract or simulated noise models. 
While these models are useful for isolating algorithmic effects, real analog hardware often exhibits more complex behavior. 
In particular, environmental conditions such as temperature can influence not only the magnitude of noise, but also the operating characteristics of the analog circuitry.
Thus, it remains unclear whether temperature-induced degradation in analog neural network inference can be modeled simply as a change in noise variance, or whether additional effects must be considered.
This distinction is important. 
If temperature only changes the variance of an otherwise stable noise process, then methods such as VANT provide a natural and direct solution. 
However, if temperature also changes signal gain, layer-wise behavior, or the effective scaling of intermediate activations, then robustness methods based only on noise variance may be incomplete.
Initial experiments applying VANT to BrainScaleS-2 (BSS-2), a mixed-signal analog neuromorphic processor, pointed in this direction.

Thus, the objective of this work is to characterize how temperature affects analog neural network inference on BSS-2 and to analyze the implications of these effects for robust training methods such as noisy training and VANT. 
Specifically, this paper makes the following contributions:
\begin{enumerate}
    \vspace{-0.2em}
    \item We characterize temperature-dependent analog neural network inference on BSS-2 with a focus on disentangling noise and systematic non-idealities.
	\item We show that temperature-induced degradation is not fully captured by noise variance alone, as temperature affects both stochastic noise and systematic non-idealities.
	\item We evaluate the effectiveness of hardware-in-the-loop training, noisy training, VANT, and different calibration strategies for mitigating temperature-induced degradation.
\end{enumerate}

\section{Background}
While analog computing can be realized with a variety of technologies, including optical \cite{optical}, photonic \cite{shen2017deep}, and phase-change memory approaches \cite{ortner2025rapid}, the most widely adopted and the focus of this work are electronic CMOS-based implementations \cite{9197673}.
In these systems, operations such as multiplication and accumulation are directly mapped to the underlying circuit dynamics, enabling highly parallel and energy-efficient computation.
Despite these advantages, analog CMOS systems remain susceptible to environmental variations, particularly temperature fluctuations. 
Although temperature compensation techniques for analog DNN accelerators have been explored \cite{11073135,9647971}, existing approaches cannot be directly applied to BSS-2 due to its distinct operating principles.

\begin{wrapfigure}{r}{0.38\textwidth}
	\centering
	\vspace{-6.2em} 
	\includegraphics[width=0.39\textwidth]{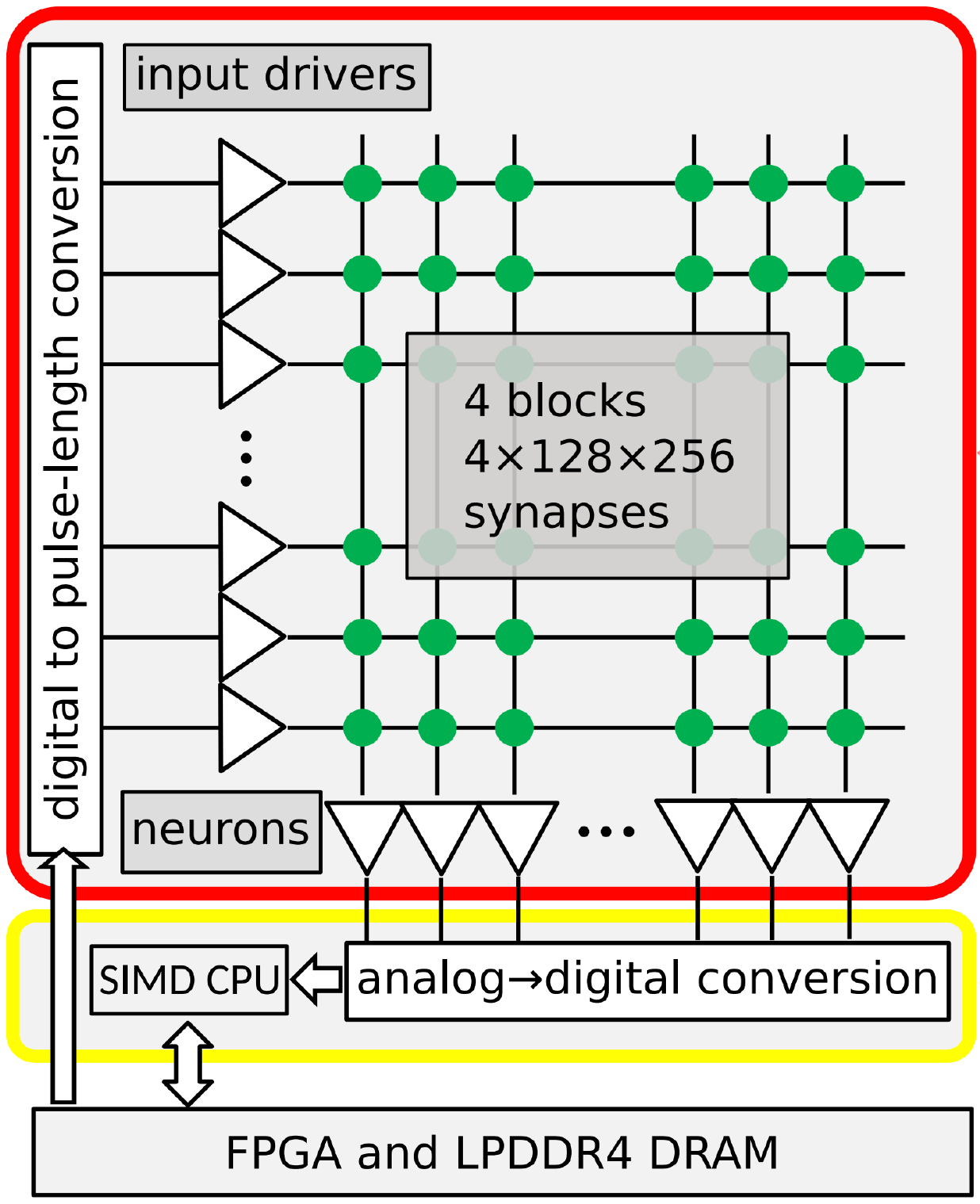}
	\caption{Internal structure of the BSS-2 ASIC, with input drivers (triangles on the left), neurons (triangles at the bottom), and synapses (green dots) \cite{bss2-figs}.}
	\vspace{-2.5em} 
	\label{fig:bss2-f1}
\end{wrapfigure}

\subsection{BrainScaleS-2}
\label{sec:bss2}
BSS-2 is a mixed-signal analog neuromorphic system based on a custom ASIC fabricated using a 65\,nm CMOS process~\cite{BSS-2}.
Although originally developed for the emulation of spiking neural networks (SNNs), the platform can also accelerate artificial neural networks (ANNs) in a non-spiking configuration by performing analog matrix–vector multiplications~\cite{weis-ANN}.

The BSS-2 ASIC comprises four blocks of $128\times256$ synapses and 512 neuron circuits, each equipped with a dedicated ADC channel for activation readout (Fig.~\ref{fig:bss2-f1}).
A single execution supports matrix–vector multiplications with matrices of up to $128\times512$ elements in signed mode or $256\times512$ elements in unsigned mode.
Larger computations must be partitioned across multiple hardware executions.

\begin{figure}
	\centering
	\includegraphics[width=0.975\textwidth]{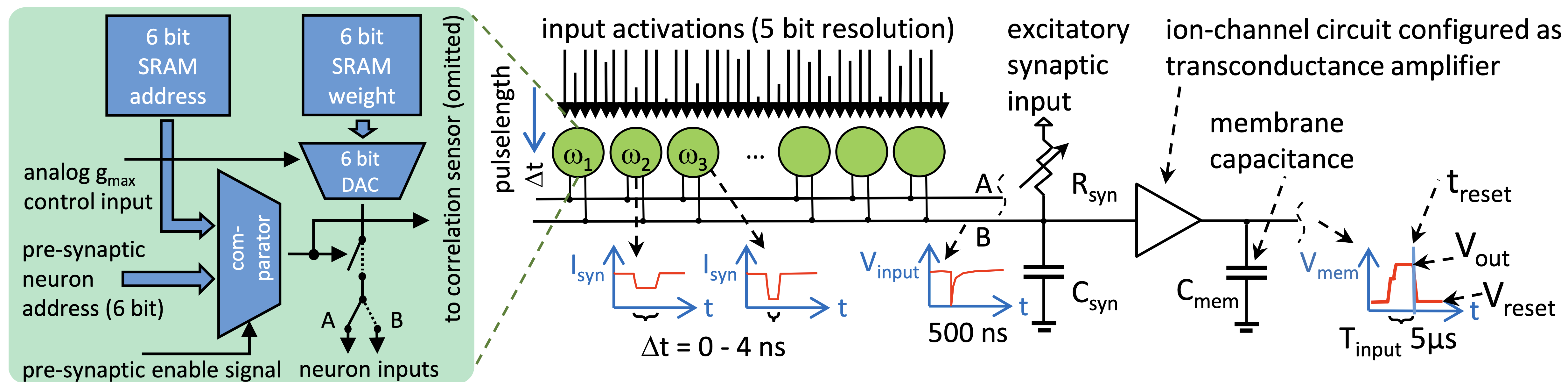}
	\caption{Principle of analog matrix–vector multiplication on BSS-2~\cite{bss2-figs}.}
	\vspace{-2em}
	\label{fig:bss2-f2}
\end{figure}

Fig.~\ref{fig:bss2-f2} illustrates the principle of analog computation on BSS-2.
Matrix–vector multiplication is realized through charge accumulation, where the product of an input value and a synaptic weight is represented as an electrical charge $Q = I \cdot \Delta t$.
The current $I$ is determined by the locally stored 6-bit unsigned synaptic weight.
In signed mode, the weight is extended by a sign bit, yielding an effective 7-bit signed representation.
The pulse duration $\Delta t$ encodes the unsigned 5-bit input value.
Each neuron accumulates the charge contributions generated by all synapses in its column.
The accumulated charge is converted into a current by an operational transconductance amplifier (OTA), integrated on the neuron’s membrane capacitance, and subsequently digitized by an 8-bit ADC.

The analog realization in BSS-2 introduces both systematic and stochastic non-idealities.
Manufacturing-induced mismatches lead to fixed-pattern variations in the electrical properties of individual circuit components and, consequently, systematic deviations between computational units.
To compensate for these variations, BSS-2 provides a set of digitally configurable parameters that are optimized during a calibration procedure.
Most importantly, the calibration adjusts the synaptic current strengths to achieve consistent responses across neurons.
Furthermore, it calibrates the pulse-generation circuits to compensate for random offsets in current-pulse encoding~\cite{weis-ANN}.
To reduce the impact of stochastic noise, BSS-2 supports resending the same input vector multiple times within a single integration phase.
Nevertheless, residual systematic and stochastic non-idealities remain after calibration.
In later sections, we revisit calibration in the context of temperature-induced variations and its impact on inference accuracy.

\subsection{Hardening Neural Networks Against Analog Non-Idealities}
\label{sec:vant}

To mitigate hardware-induced non-idealities, prior work has shown that hardware-in-the-loop training is highly effective~\cite{weis-ANN,bernhard-incremental}.
By incorporating the target hardware into the forward path during training, the model can directly adapt to hardware-specific imperfections, similar to quantization-aware training~\cite{jmlr2024}.
However, its computational cost and dependence on a specific hardware state limit its applicability in dynamic environments.

An alternative is noisy training, which improves robustness by injecting additive Gaussian noise during training~\cite{noisy_machine}.
Most noisy training approaches assume a fixed noise level and therefore do not account for variations caused by changing operating conditions, such as temperature fluctuations, voltage instability, or device aging.
Variance-Aware Noisy Training (VANT)~\cite{xiaowang2025} addresses this limitation by sampling the injected noise from a distribution rather than using a fixed value.
Specifically, the noise level is drawn from $\mathcal{N}(\alpha \cdot \sigma_\text{train}, \theta)$, where $\sigma_\text{train}$ denotes the nominal noise level of the target hardware and $\alpha$ and $\theta$ control the mean and variance of the sampled noise.
This enables the model to become robust to a range of noise conditions rather than a single operating point, making VANT a promising approach for deployment under varying operating conditions.

\section{Noise Characterization of BrainScaleS-2 (BSS-2)}
\label{sec:characterization}
In this section, we present a detailed noise characterization using the BSS-2 system.
We first examine the per-neuron behavior at room temperature and then investigate the effects of temperature on neuron-wise and global system characteristics.
All experiments presented in this and the following sections were performed on the same device, operating in the signed weight mode.

\subsection{Metrics}\label{sec:hw-measurement}

To evaluate hardware performance in matrix--vector multiplication tasks, we conduct a series of experiments to characterize the associated non-idealities. 
For this purpose, 100 matrix--vector multiplications are executed on the hardware platform and compared against numerically exact reference computations.

Input vectors consist of 128 elements, while the matrices have dimensions of $128 \times 256$. Both sizes are chosen such that they map efficiently to the hardware.
All matrix and vector elements were independently sampled from uniform distributions spanning the full value ranges supported by BSS-2.
This preserves realistic activity patterns by assigning different computations across neurons rather than broadcasting an identical computation, which could induce atypical crosstalk.
For each of the 100 executions, the matrix and input vector were resampled, reducing bias from any particular neuron-computation assignment and enabling a fair statistical comparison across neurons.

For the output vector, the hardware introduces a scaling factor relative to the exact result, referred to as the gain $G_{\mathrm{op}}$.
It is defined as the median of element-wise ratios between the hardware output $\mathbf{y}_{\mathrm{hw}}$ and the exact output vector $\mathbf{y}_{\mathrm{exact}}$, where $i \in \{1,\dots,K\}$ indexes vector components:

\vspace{-0.5em}
\[
G_{\mathrm{op}} =
\operatorname{median} \left( \frac{y_{\mathrm{hw},i}}{y_{\mathrm{exact},i}} \right).
\]

The hardware error $\Delta_i$ is then defined as the difference between the hardware output and the gain-scaled exact output:
\vspace{-0.3em}
\[
\Delta_i = y_{\mathrm{hw},i} - G_{\mathrm{op}} \cdot y_{\mathrm{exact},i}.
\]

After $N$ matrix--vector multiplication runs, we compute the neuron-wise mean error $\mu_i$ and standard deviation $\sigma_i$ of the error:
\vspace{-0.3em}
\[
\mu_i = \frac{1}{N}\sum_{j=1}^{N} \Delta_{i,j},
\qquad
\sigma_i = \sqrt{\frac{1}{N}\sum_{j=1}^{N} \left(\Delta_{i,j} - \mu_i\right)^2}.
\]

\begin{figure}[t]
	\centering
	
	\begin{subfigure}[t]{0.32\linewidth}
		\centering
		\includegraphics[width=\linewidth]{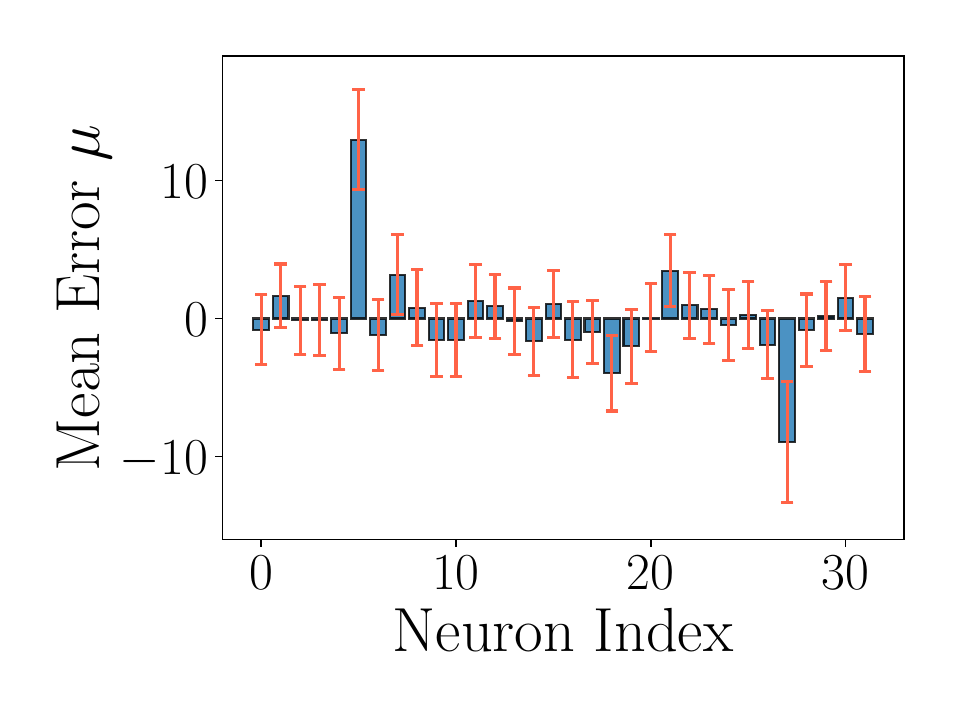}
		\caption{Mean and standard deviation of per-neuron error for a subset of 32 neurons ($T_{\mathrm{operating}}=40^\circ\mathrm{C}$). }
		\label{fig:error-per-neuron}
	\end{subfigure}
	\hfill
	\begin{subfigure}[t]{0.32\linewidth}
		\centering
		\includegraphics[width=\linewidth]{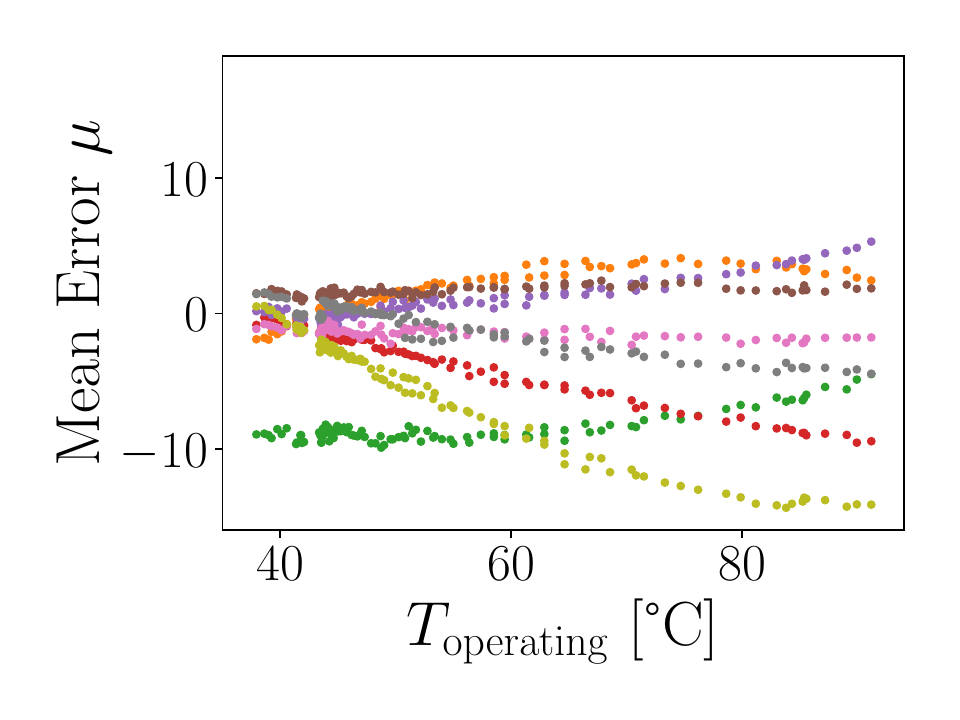}
		\caption{Per-neuron mean error over temperature for a subset of 8 neurons (each color represents one neuron).}
		\label{fig:per-neuron-error-over-temp}
	\end{subfigure}
	\hfill
	\begin{subfigure}[t]{0.32\linewidth}
		\centering
		\includegraphics[width=\linewidth]{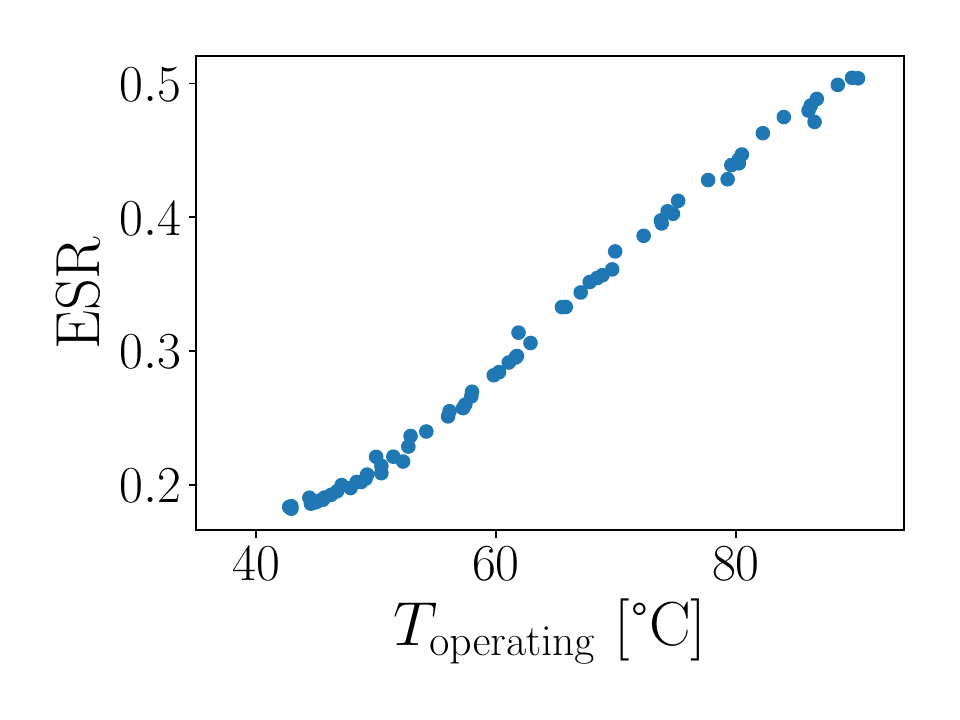}
		\caption{Error-to-signal ratio over temperature (described in section~\ref{sec:temperature-noise}).}
		\label{fig:ser-temp}
	\end{subfigure}
	
    \caption{Error analysis over 100 random matrix–vector multiplications on BSS-2, reflecting the combined behavior of noise and systematic non-idealities.}
	\vspace{-2em} 
	\label{fig:per-neuron-analysis}
\end{figure}

Fig.~\ref{fig:error-per-neuron} shows $\mu_i$ and $\sigma_i$ for 32 representative neurons. 
The mean error $\mu_i$ (blue bars) varies substantially across neurons and reflects deterministic, device-dependent effects associated with calibration and hardware-specific imperfections. 
We refer to these contributions as \textit{systematic non-idealities}.

In contrast, $\sigma_i$ (red error bars) is comparatively uniform across neurons, indicating a largely consistent stochastic contribution, which we refer to as \textit{noise}.

Across many neurons, $\sigma_i$ is typically dominant, indicating that noise is the primary error source. 
However, for a subset of neurons, we find that $|\mu_i| \gg \sigma_i$, indicating that systematic non-idealities can locally dominate and lead to significant yet structured deviations.
This motivates the following investigation of the respective impacts of noise and systematic non-idealities on computations.

\subsection{Temperature-Dependent Noise Behavior}
\label{sec:temperature-noise}

\begin{wrapfigure}{r}{0.4\textwidth}
	\centering
	\vspace{-2.2em} 
	\includegraphics[trim={15cm 0 35cm 0},clip,width=0.38\textwidth]{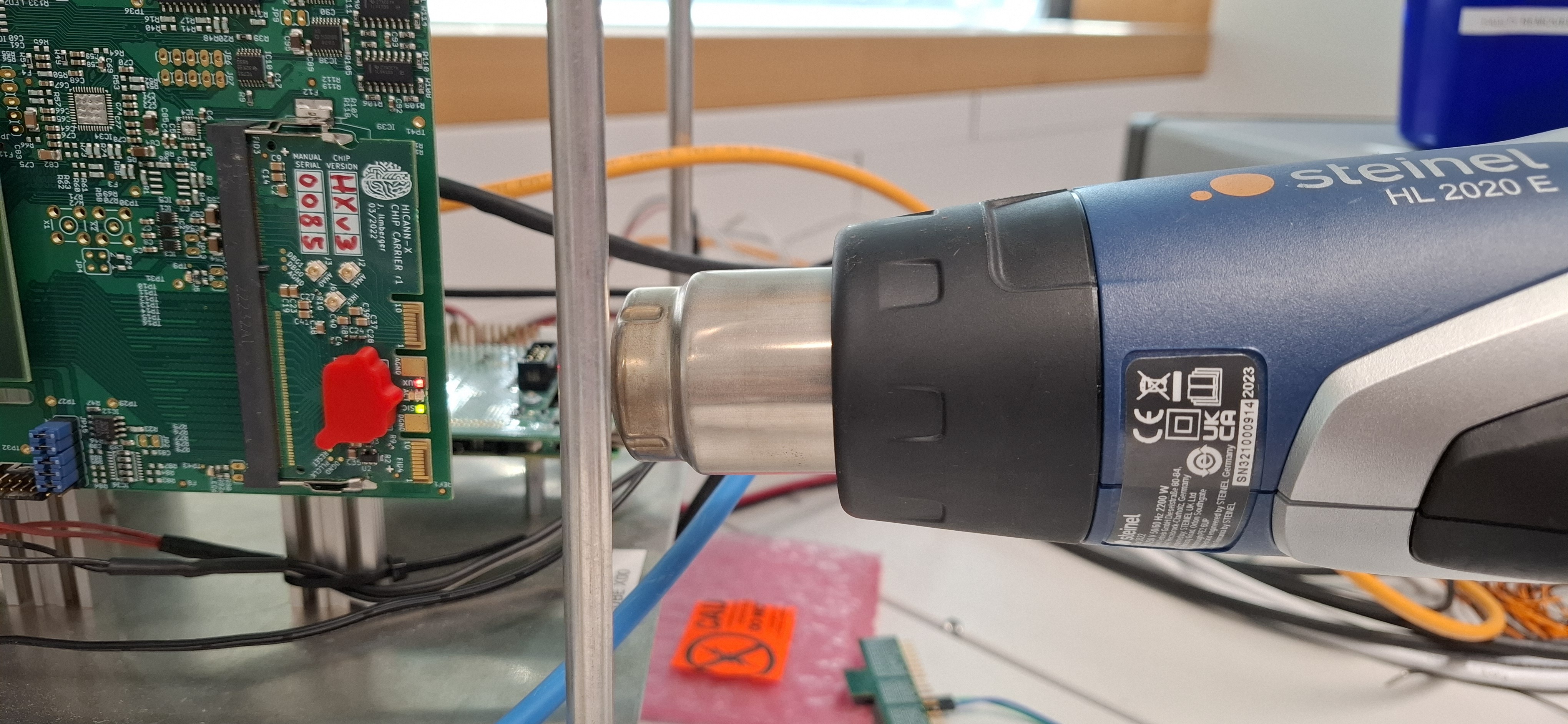}
	\caption{Experiment setup for temperature measurements with BSS-2.}
	\vspace{-2.5em} 
	\label{fig:exp_setup}
\end{wrapfigure}

In addition to the gain and error measurements described in Section~\ref{sec:hw-measurement}, we investigate how these quantities evolve under varying temperature conditions. 
Temperature effects on lower-level BSS-2 circuit characteristics have previously been studied~\cite{transistors}; however, their impact on the presented gain and error metrics remains unclear.

To vary the operating temperature without hardware modifications, a heat gun was used (Fig.~\ref{fig:exp_setup}), while the device temperature was monitored using the on-board sensor.
Due to ambient conditions and self-heating, the baseline operating temperature was approximately $40^\circ\mathrm{C}$.
The temperature was increased to about $90^\circ\mathrm{C}$ over 6 min and then allowed to cool back to baseline over 10 min, covering an effective range of $50^\circ\mathrm{C}$ in a total of 16 min.
Measurements were recorded every 10 s, yielding nearly 100 measurement points.

As a first step, the impact of temperature changes on both noise and systematic non-idealities is analyzed.
Fig.~\ref{fig:per-neuron-error-over-temp} illustrates the temperature-dependent evolution of the non-idealities for a subset of 8 neurons. 
At baseline temperature, most neurons exhibit near-optimal behavior, with mean errors close to zero.
In contrast, the green dots correspond to an outlier neuron exhibiting a substantial mean error.
With increasing temperature, systematic non-idealities shift with varying magnitude and direction across neurons, revealing heterogeneous temperature effects.
While most neurons shift further from zero, indicating degraded behavior, the green outlier partially recovers.

To evaluate the impact of these deviations on signal quality, we computed the error-to-signal ratio (ESR), defined as the ratio of  the absolute error to the absolute signal amplitude, where the error comprises both stochastic noise and systematic non-idealities. 
As shown in Fig.~\ref{fig:ser-temp}, the mean ESR averaged across all neurons increases with temperature, indicating that elevated temperatures degrade signal fidelity.

\begin{figure}[t]
    \centering

    \begin{subfigure}[t]{0.32\textwidth}
        \centering
        \includegraphics[width=\linewidth]{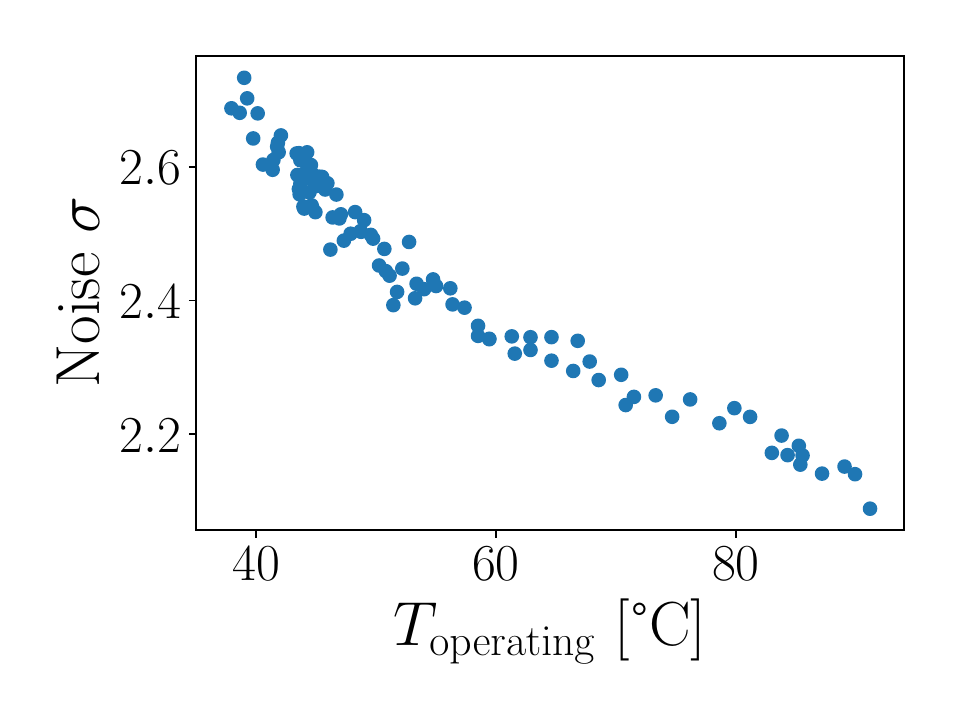}
        \caption{Noise}
        \label{fig:noise-std}
    \end{subfigure}
    \begin{subfigure}[t]{0.32\textwidth}
        \centering
        \includegraphics[width=\linewidth]{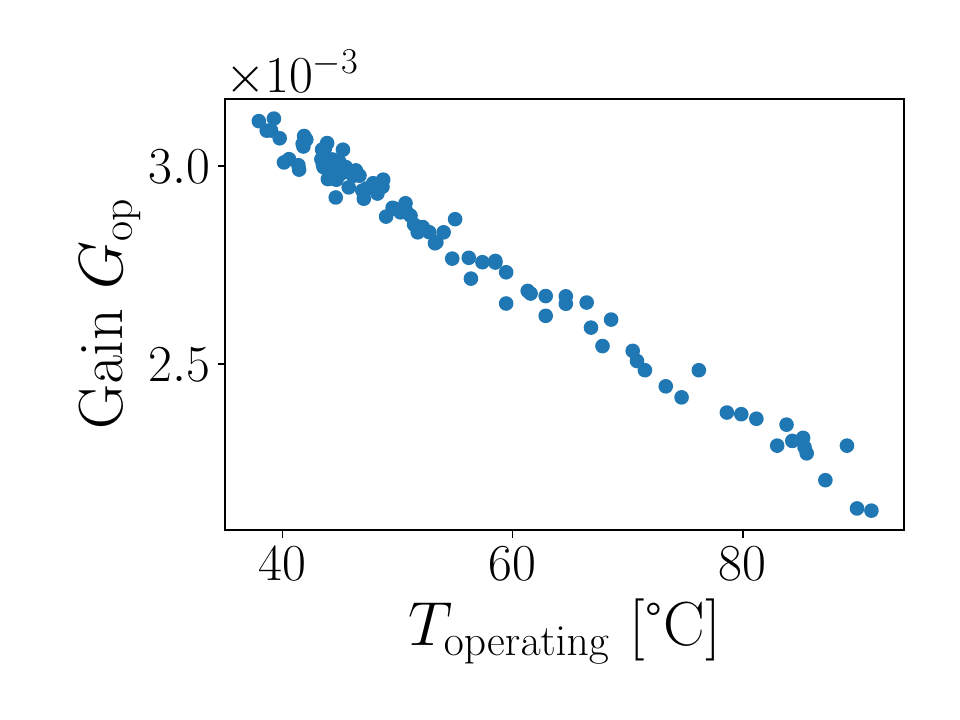}
        \caption{Gain}
        \label{fig:gain-temp}
    \end{subfigure}
    \begin{subfigure}[t]{0.32\textwidth}
        \centering
        \includegraphics[width=\linewidth]{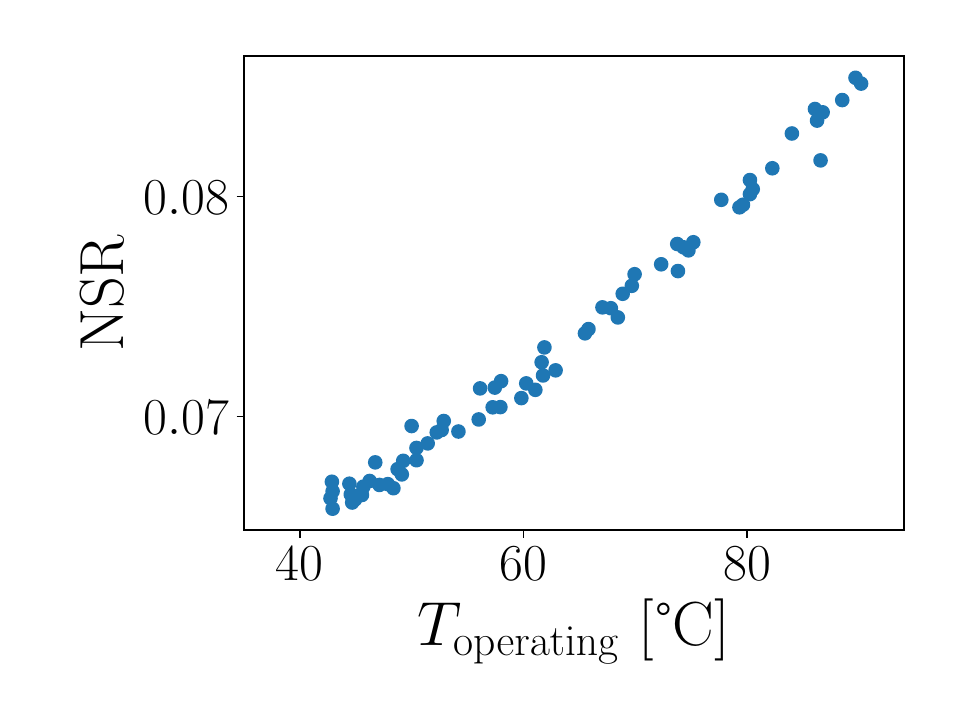}
        \caption{Noise-to-signal ratio}
        \label{fig:snr-temp}
    \end{subfigure}

    \caption{Temperature-dependent behavior of noise (left) and gain (middle) and noise-to-signal ratio (right).}
    \vspace{-2em} 
    \label{fig:temp-analysis}
\end{figure}

We assume that both stochastic noise and systematic non-idealities contribute additively to the mean ESR, which allows us to disentangle their contributions.
To this end, we first investigate the evolution of the noise and gain with increasing temperature.
Fig.~\ref{fig:noise-std} shows the mean standard deviation $\mathbf{\sigma}$, averaged across all neurons, while Fig.~\ref{fig:gain-temp} depicts the corresponding gain $G_{\mathrm{op}}$. 
Both quantities decrease with increasing temperature.

Despite this reduction, the effective noise-to-signal ratio (NSR), shown in Fig.~\ref{fig:snr-temp}, increases with temperature.
Importantly, the NSR captures only stochastic noise and excludes any changes in systematic non-idealities.
When comparing the ESR results (Fig.~\ref{fig:ser-temp}) with the NSR results (Fig. ~\ref{fig:snr-temp}), it becomes apparent that while both measurements trend in the same direction,
the stochastic noise contribution to degraded signal quality at elevated temperatures is small and systematic non-idealities dominate.

\section{Error Countermeasures}
To further investigate how temperature-induced variations uncovered in Section \ref{sec:characterization} affect neural network performance, we train and deploy neural networks directly on the real hardware platform.
This setup allows us to capture the full impact of hardware-in-the-loop dynamics under realistic operating conditions. 
To evaluate the effectiveness of different countermeasures, we conduct experiments across a range of operating temperatures, systematically assessing their ability to mitigate performance degradation under varying thermal noise conditions.
\subsection{Dataset and Neural Network Architecture}

Experiments were conducted on a ten-class classification task derived from the Google Speech Commands (GSC) dataset~\cite{GSC}. 
The audio samples are pre-processed to a compact yet informative mel-spectrogram representation in the form of a $64 \times 40$ feature matrix, making it well suited for deployment on a small-scale MLP architecture. 

The neural network consists of three fully-connected layers. 
Since layers exceeding the hardware's native computational capacity require multiple sequential executions, noise accumulates beyond single-pass estimates in these layers. 
Moreover, the observed noise depends not only on hardware properties but also on the concrete distribution of activations and weights, necessitating a layer-wise noise estimation. 
Table~\ref{tab:model-architecture} summarizes the network architecture together with the expected noise standard deviation during execution on BSS-2 hardware, using samples from the target dataset together with weights from a quantization-aware pretrained model. 
The table further reports the resend counts and resulting gain factors. 
Bias terms are disabled, as they are unsupported on BSS-2.
\begin{table}[t]
\centering
\caption{Layer-wise network architecture and BSS-2 execution parameters.}
\label{tab:model-architecture}

\renewcommand{\arraystretch}{1.2}
\begin{tabular*}{\linewidth}{*{7}{>{\centering\arraybackslash}p{1.6cm}}}
\toprule
Layer &
Input Features &
Output Features &
Bias &
No. of Resends &
Gain &
Noise Std. 
\\
\midrule
1 & 2560 & 128 & No & 2 & 0.003 & 5.4 \\
2 & 128  & 128 & No & 4 & 0.006 & 1.5 \\
3 & 128  & 10  & No & 4 & 0.006 & 1.7 \\
\bottomrule
\end{tabular*}
\vspace{-1.5em} 
\end{table}

\subsection{Accuracy and Robustness Evaluation}

We investigate the impact of different countermeasures on model accuracy and robustness. 
The evaluated methods can be grouped into two categories: \textit{simulation-based} and \textit{hardware-based} approaches.  
All models were trained using the Adam optimizer with a batch size of 128. 
The learning rate was initialized at $1 \times 10^{-3}$ and gradually reduced to $1 \times 10^{-6}$ using a cosine annealing schedule.

\vspace{-0.8em} 
\subsubsection{Simulation-Based Training}
For the simulation-based methods, we aimed to reproduce hardware effects entirely in software. 
To this end, a custom PyTorch layer was implemented to emulate execution on the BSS-2 system. 
The layer incorporates the hardware-specific quantization described in Section~\ref{sec:bss2} with signed weights, injects noise, and scales outputs according to gain factors obtained from hardware characterization measurements. 
During backpropagation, quantization and noise injection are bypassed using a straight-through estimator (STE)~\cite{jmlr2024}, while only the scaling factors are retained and reapplied in the backward pass.
All simulation-based methods were trained in full precision for the first ten epochs. 
Quantization-aware training (QAT)~\cite{jmlr2024} and scaling effects were enabled from epoch 11 onward. 
The evaluated training strategies differ only in how noise is incorporated during training:

\begin{itemize}
	\item \emph{Noise-Free QAT:}
		No noise was injected during training. 
		Quantization and scaling effects were simulated from epoch 11 onward, and training continued for a total of 300 epochs.
	\item \emph{Standard noisy training:}
		Noise injection was enabled from epoch 21 onward, with the noise level gradually increasing until it reaches the target strength listed in Table~\ref{tab:model-architecture} at epoch 121. 
		Training then continued under full-noise conditions until epoch 300.
	\item \emph{Variance-Aware Noisy Training (VANT):}
		The noise schedule was identical to that of standard noisy training. 
		However, the target noise strength was sampled individually for each training sample.  
		Following~\cite{xiaowang2025}, we set $\sigma_{\text{train}}$ equal to the target noise strength and express $\theta$ as a scaling factor of $\sigma_{\text{train}}$ to account for layer-dependent effective noise levels.
\end{itemize}

\vspace{-1em} 
\subsubsection{Hardware-in-the-Loop Training}
In addition to simulation-based approaches, we performed hardware-in-the-loop training using the BSS-2 hardware during the forward pass. 
Since training entirely with hardware-in-the-loop execution resulted in prohibitively slow convergence, models were initialized from a network pretrained with standard noisy training and subsequently fine-tuned for 80 epochs using hardware execution in the forward path.

\vspace{-0.8em} 
\subsubsection{Hardware Calibration}

To assess the influence of calibration temperature, calibration settings were generated at $T_{\mathrm{cal}} \in \{ 40^\circ\mathrm{C},60^\circ\mathrm{C},80^\circ\mathrm{C} \}$,
yielding three corresponding hardware-in-the-loop trained models.
Hardware-in-the-loop training itself was performed at the baseline operating temperature of $40^\circ\mathrm{C}$, since training at elevated temperatures is impractical due to the associated time cost.

\vspace{-0.8em} 
\subsubsection{Temperature Robustness Evaluation}

\begin{wrapfigure}{r}{0.4\textwidth}
	\centering
	\vspace{-3.0em} 
	\includegraphics[width=0.38\textwidth]{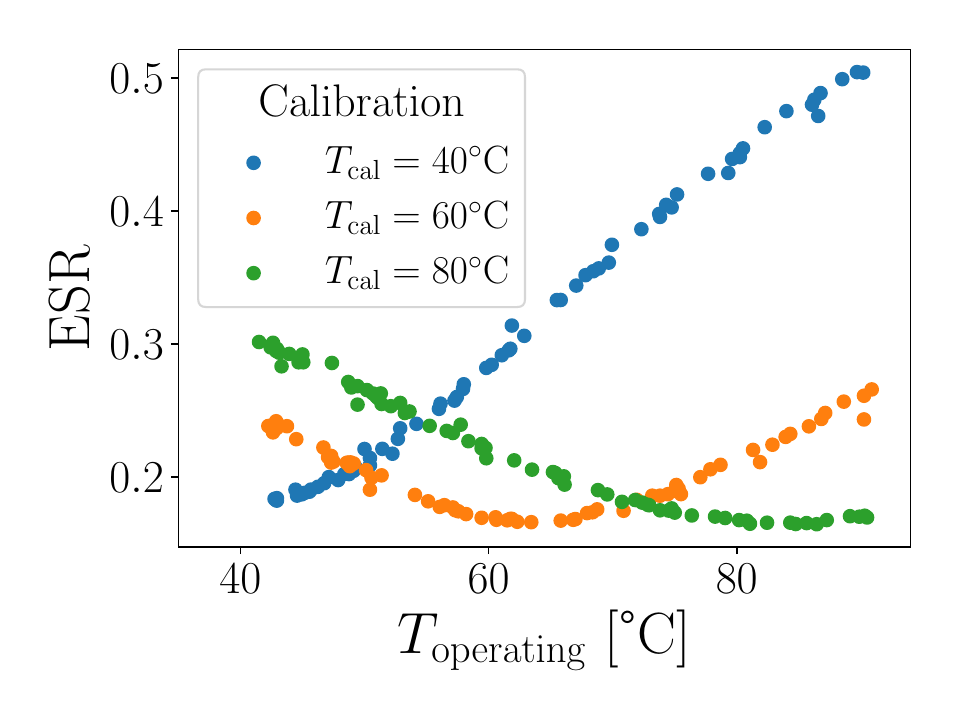}
	\caption{Error-to-signal ratio over operating temperature $T_{\mathrm{operating}}$ for each calibration.}
	\vspace{-2em} 
	\label{fig:esr_all_calibs}
\end{wrapfigure}

\begin{figure}[tb]
	\centering
	
	\begin{subfigure}[t]{0.32\linewidth}
		\centering
		\includegraphics[width=\linewidth]{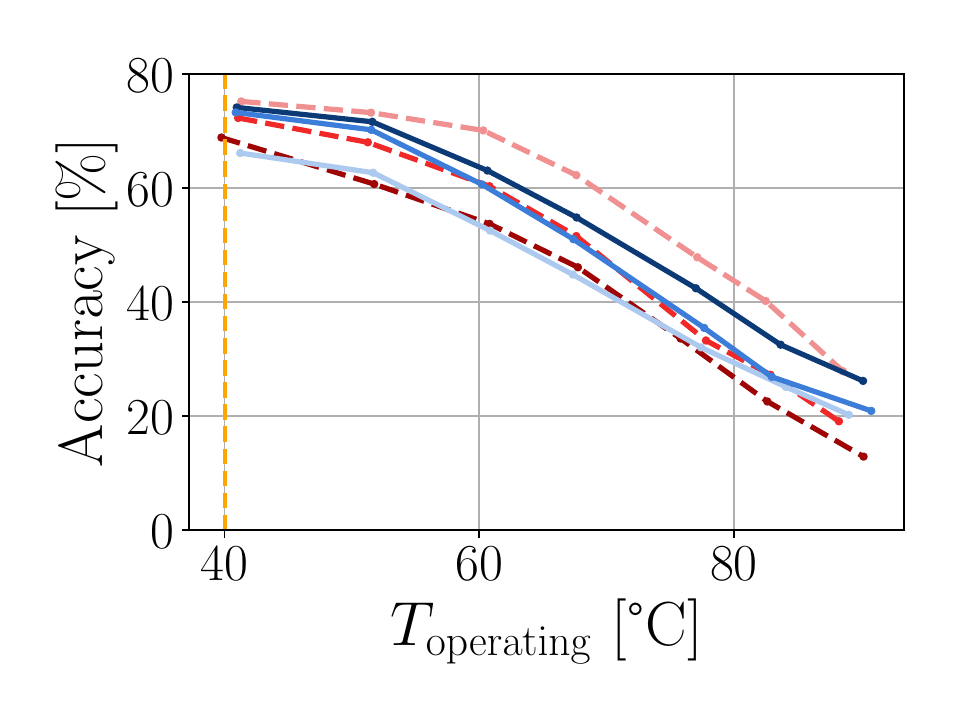}
		\caption{Inference $T_{\mathrm{cal}} = 40^\circ\mathrm{C}$}
		\label{fig:acc-temp-40}
	\end{subfigure}
	\hfill
	\begin{subfigure}[t]{0.32\linewidth}
		\centering
		\includegraphics[width=\linewidth]{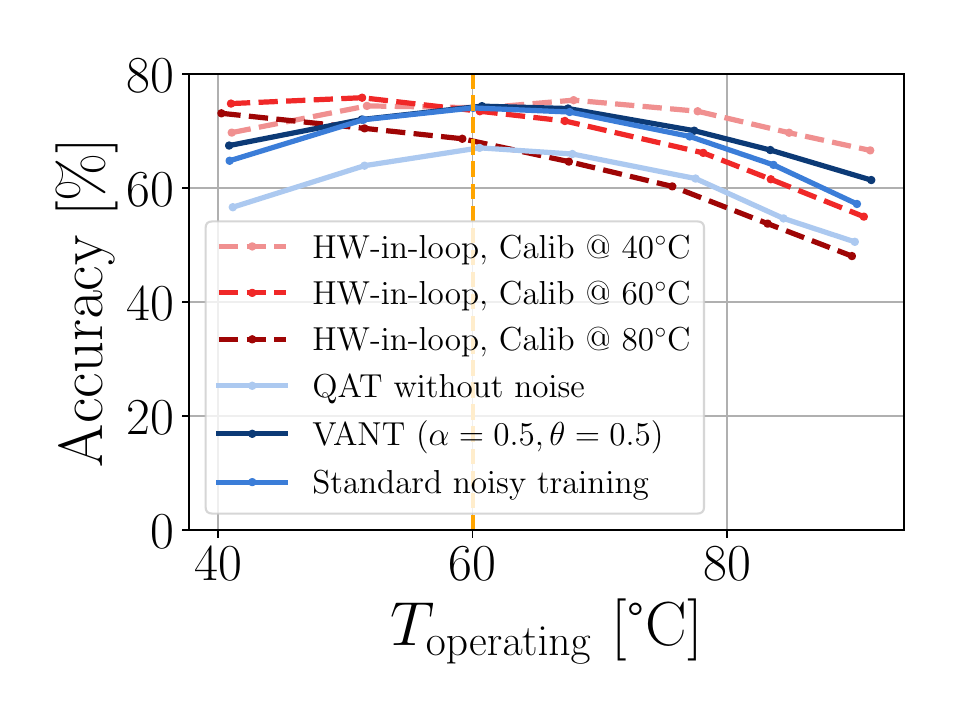}
		\caption{Inference $T_{\mathrm{cal}} = 60^\circ\mathrm{C}$}
		\label{fig:acc-temp-60}
	\end{subfigure}
	\hfill
	\begin{subfigure}[t]{0.32\linewidth}
		\centering
		\includegraphics[width=\linewidth]{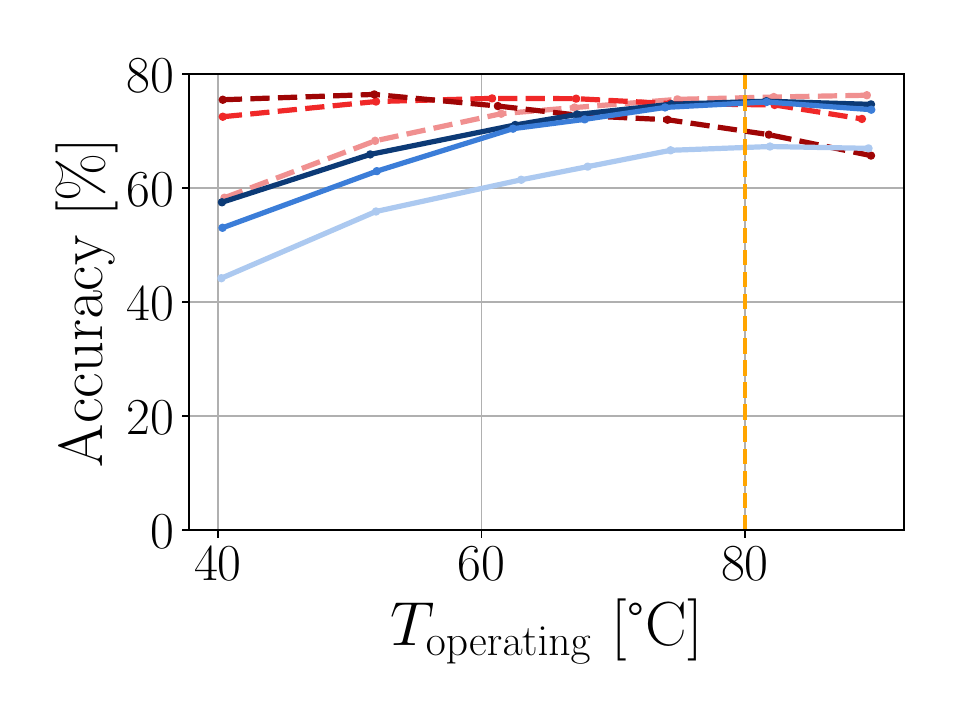}
		\caption{Inference $T_{\mathrm{cal}} = 80^\circ\mathrm{C}$}
		\label{fig:acc-temp-80}
	\end{subfigure}
	
	\caption{Validation accuracy versus operating temperature for different training approaches under three hardware calibration conditions. Each subplot represents inference using one calibration created at $T_{\mathrm{cal}}$.}
	 \vspace{-2em} 
	\label{fig:acc-temp}
	
\end{figure}

Robustness to temperature variations was evaluated by performing inference at operating temperatures ($T_{\mathrm{operating}}$) ranging from $40^\circ\mathrm{C}$ to $90^\circ\mathrm{C}$. 
Fig.~\ref{fig:acc-temp} shows the corresponding validation accuracy curves. 

For all simulation-based training models, peak accuracy is observed near $T_{\mathrm{operating}} = T_{\mathrm{cal}}$, indicating that calibration is most effective close to the temperature at which it was generated. 
This trend is consistent with the ESR measurements in Fig.~\ref{fig:esr_all_calibs}, where the minimum ESR occurs near the corresponding calibration temperature. 
Compared with noise-free QAT, both standard noisy training and VANT substantially improve robustness, with VANT providing an additional, albeit modest, benefit when the operating temperature deviates from the calibration point.

In contrast, the hardware-in-the-loop trained models exhibit markedly different behavior.
Unlike the simulation-based approaches, the highest accuracies are not necessarily observed when $T_{\mathrm{operating}} = T_{\mathrm{cal}}$.
Instead, the observed behavior depends on the calibration used during training, which may differ from the calibration applied during inference.

Models trained with calibrations generated at elevated temperatures ($> 40^\circ\mathrm{C}$) tend to perform better below their calibration point, since hardware-in-the-loop training occurs at baseline temperature regardless of calibration, which enables the model to compensate for non-idealities expected in that regime. 
Above the calibration temperature, however, accuracy degrades rapidly as the training setting no longer reflects actual operating behavior.

As a result, using an inference calibration generated at the higher end of the temperature range, as shown in Fig.~\ref{fig:acc-temp-80}, can provide high robustness across the entire investigated temperature range, as most operating temperature settings fall below the calibration point.
In particular, the model trained using the $60^\circ\mathrm{C}$ calibration demonstrates remarkable robustness.

Overall, we observe that in many cases the use of VANT is recommended as a first mitigation strategy for temperature-induced effects,
as it delivers reasonable performance at no hardware training overhead.
For further improvements, HW-in-the-loop training on a well-selected calibration can achieve better performance.

\section{Summary and Outlook}

This work investigates the impact of temperature on analog neural-network inference using BrainScaleS-2. 
The observed temperature-dependent effects manifest as gain drift, neuron-specific deviations, and changes in the signal-to-noise ratio. Together, these effects lead to measurable accuracy degradation that cannot be attributed to noise variations alone. 
We analyze the error-to-signal and noise-to-signal ratios to separate stochastic and systematic contributions.
The analysis suggests that systematic non-idealities are the dominant source of degradation.

In exploring mitigation strategies for temperature-induced accuracy losses, we obtain three main findings. First, hardware-in-the-loop training generally produces the best-performing models, highlighting the importance of calibration awareness during both training and execution. 
Second, through the first application of VANT to real neuromorphic hardware, 
we observe that VANT generally outperforms standard noisy training, with the clearest benefits occurring when the operating temperature deviates from the calibration temperature.
Moreover, under certain conditions, simulation-based training with VANT can achieve performance comparable to hardware-in-the-loop training.
Third, recalibrating the chip around its expected operating temperature can shift the optimum operating point and significantly improve robustness. This trend appears to hold across all investigated training methods.

Looking ahead, one promising direction is the use of more expressive models, which have been shown to exhibit improved robustness in the context of VANT.
Furthermore, systematic non-idealities exhibit a highly predictable drift as a function of temperature. 
We therefore propose that explicitly modeling the temperature dependence of these non-idealities could enable a substantially improved training procedure. 
Although such an approach requires per-chip calibration, it may provide a path toward deep neural networks that are better adapted to the temperature variations encountered in real-world deployment scenarios.

\section*{Acknowledgments}
We gratefully acknowledge the members of the Electronic Visions group for the design and provision of the chip used in this work, as well as for their technical support and access to the experimental infrastructure. We especially thank Johannes Schemmel for supporting this collaboration, and Yannik Stradmann for serving as our primary point of contact.

\begin{credits}
	\subsubsection{\discintname}
	The authors have no competing interests to declare that are
	relevant to the content of this article.
\end{credits}

\bibliographystyle{splncs04}
\bibliography{references}

\end{document}